\documentclass{article}

\usepackage{microtype}
\usepackage{graphicx}
\usepackage{subfigure}
\usepackage{booktabs} % for professional tables
\usepackage{xcolor}

\usepackage{hyperref}
\usepackage{url}

\usepackage[final]{changes} %Use this to hide all comments.
\definechangesauthor[name={X.~X.}, color={orange}]{xx}
\definechangesauthor[name={Y.~Y.}, color={blue}]{yy}
\definechangesauthor[name={K.~K.}, color={green}]{kk}
\definechangesauthor[name={B.~L.}, color={orange}]{bl}

\usepackage[accepted]{tccmlicml2021}

\icmltitlerunning{Tackling the Overestimation of Forest Carbon with DL and Aerial Imagery}

\begin{document}

\twocolumn[
\icmltitle{Tackling the Overestimation of Forest Carbon \\ with Deep Learning on Aerial Imagery}



\begin{icmlauthorlist}
\icmlauthor{Gyri Reiersen}{tum,eth}
\icmlauthor{David Dao}{eth}
\icmlauthor{Björn Lütjens}{mit}
\icmlauthor{Konstantin Klemmer}{tum}
\icmlauthor{Xiaoxiang Zhu}{tum}
\icmlauthor{Ce Zhang}{eth}

\end{icmlauthorlist}

\icmlaffiliation{tum}{Department of Informatics, Technical University of Munich, Munich, Germany}
\icmlaffiliation{eth}{Department of Computer Science, ETH Zurich, Zurich, Switzerland}
\icmlaffiliation{mit}{Department of Aeronautics and Astronautic, Massachusetts Institute of Technology, Cambridge, USA}

\icmlcorrespondingauthor{Gyri Reiersen}{gyri.reiersen@tum.de}
\icmlcorrespondingauthor{David Dao }{david.dao@inf.ethz.ch}

\icmlkeywords{Machine Learning, ICML, Deep Learning, Remote Sensing, Forest Carbon Stock, Carbon Offsets}

\vskip 0.3in
]



\printAffiliationsAndNotice{\icmlEqualContribution} % otherwise use the standard text.

\begin{abstract}
Forest carbon offsets are increasingly popular and can play a significant role in financing climate mitigation,  forest conservation, and reforestation.
Measuring how much carbon is stored in forests is, however, still largely done via expensive, time-consuming, and sometimes unaccountable field measurements. To overcome these limitations, many verification bodies are leveraging machine learning (ML) algorithms to estimate forest carbon from satellite or aerial imagery.
Aerial imagery allows for tree species or family classification, which improves on the satellite imagery-based forest type classification. However, aerial imagery is significantly more expensive to collect and it is unclear by how much the higher resolution improves the forest carbon estimation. 
In this proposal paper, we describe the first systematic comparison of forest carbon estimation from aerial imagery, satellite imagery, and ``ground-truth`` field measurements via deep learning-based algorithms for a tropical reforestation project. Our initial results show that forest carbon estimates from satellite imagery can overestimate aboveground biomass by up to 10-times for tropical reforestation projects.
The significant difference between aerial and satellite-derived forest carbon measurements shows the potential for aerial imagery-based ML algorithms and raises the importance to extend this study to a global benchmark between options for carbon measurements.
\end{abstract}

\section{Introduction}
\label{intro}

The deterioration of the natural world is unparalleled in human history and a key driver of the climate crisis. Since 2000, we have lost 361 million ha of forest cover (the size of Europe)~\citep{Hansen_2013} accounting for 18\% of global anthropogenic emissions~\cite{IPCC_land_use_2019}. The causes of deforestation are mostly economically driven and major conservation efforts are underway to mitigate and safeguard against these losses.

Carbon offsets are a way of financing and trading on the capture of carbon for businesses and governments. The carbon offsetting market is expected to grow by a factor of 100 until 2050 and demand is rapidly increasing \cite{voluntaryMarket}.

Recent investigations \cite{Badgley21, West24188} have shown that the current manual practices systematically overestimate forestry carbon offsetting projects with up to 29\% of the offsets analyzed, totaling up to 30 million tCO2e and worth approximately \$410 million. There is thus a need for higher quality carbon offsetting protocols and higher transparency and accountability in the monitoring, reporting, and verification (MRV) of these projects \cite{haya2020}.

Several verification bodies and academic environments are currently developing remote sensing technologies to automate parts of the certification process of forestry carbon offsetting projects \cite{rs12111824, daogainforest}. Remote sensing through satellite or aerial imagery and lidar combined with ML models can be used to estimate carbon stock baselines and additionality, and for MRV of projects. Compared to current manual estimates, these advancements reduce time and cost and increase transparency and accountability, thus lowering the threshold for forest owners and buyers to enter the market \cite{lut2019}.

Satellite imagery is increasing in quality and availability and combined with state-of-the-art deep learning, promises to soon map every tree on earth \cite{Hanan2020} and carbon stock globally \cite{Saatchi9899}. Nevertheless, these algorithms risk additionally contributing to the systematic overestimation of carbon stocks, not reducing it.

\section{Quantifying the difference in forest carbon stock estimations and field measurements}

To quantify the difference between the estimated forest carbon stock taken from available remote sensing products, we propose a study of field measurements of six cacao agro-forestry sites in the central coastal region of Ecuador eligible for carbon offsetting certification collected in 2020. See Table \ref{project-table} for information on each site. By mapping field measurements to trees instances from drone imagery, an end-to-end deep learning-based carbon stock estimations can be done for each individual tree as seen in Figure \ref{fig:appendix} in the Appendix. Calculating the carbon stock at an individual tree level increases the accuracy of the estimations as it allows both species \cite{schiefer20} and metrics \cite{Omasa2003} to be detected.

\begin{table}[t]
\caption{Overview of the six project sites in Ecuador, as gathered in field measurements. Aboveground biomass density (AGB) is measured in metric tons per hectare and area in hectares.}

\label{project-table}
\vskip 0.15in
\begin{center}
\begin{small}
\begin{sc}
\begin{tabular}{c c c c c}
\toprule 
Site & No. of & No. of & Plot & AGB \\
no. & Trees & Species & Area & density\\

\midrule
1 & 743 & 18 & 0.53 & 19 \\ 
2 & 929 & 22 & 0.47 & 27 \\ 
3 & 846 & 16 & 0.48 & 24 \\
4 & 789 & 20 & 0.51 & 24 \\ 
5 & 484 & 12 & 0.56 & 17 \\ 
6 & 872 & 14 & 0.62 & 29 \\

\bottomrule
\end{tabular}
\end{sc}
\end{small}
\end{center}
\vskip -0.1in
\end{table}

\begin{table}[b]
\caption{Results from AGB density estimations derived from satellite-based data. Aboveground biomass (AGB) is measured in metric tonnes per hectare. The overestimation factor is calculated from comparing the ground truth to the filtered estimation.}
\label{result-table}
\vskip 0.15in
\begin{center}
\begin{small}
\begin{sc}
\begin{tabular}{ccccc}
\toprule
Site & Ground & Filtered &Over\\
no. &Truth & &estimation\\
\midrule
1 & 19 & 176 & $\times$9.2 \\ 
2 & 27 & 160 & $\times$5.9\\
3 & 24 & 47 & $\times$2.0\\
4 & 24 & 62 & $\times$2.6 \\ 
5 & 17 & 19 & $\times$1.1\\ 
6 & 29 & 141 & $\times$4.9 \\ 

\bottomrule
\end{tabular}
\end{sc}
\end{small}
\end{center}
\vskip -0.1in
\end{table}

\subsection{Data}
Field measurements were taken manually for all live trees and bushes within the site polygon and include GPS location, species, and diameter at breast height (DBH). Drone imagery was captured by an RGB camera from a Mavic 2 Pro drone in 2020. Each site is around 0.5 ha, mainly containing banana trees (Musaceae) and cocoa plants (Cocoa). The aboveground biomass (AGB) is calculated using published allometric equations for tropical agro-forestry, namely fruit trees \ref{eq1} \cite{segura2006}, banana trees \ref{eq2} \cite{van2002}, cacao \ref{eq3} \cite{Yuliasmara2009}, and shade trees (timber) \ref{eq4} \cite{Brown1992}. These are commonly used in global certification standards.

\begin{equation} \label{eq1}
log_{10}AGB_{fruit} = -0.834 + 2.223 (log_{10}DBH)
\end{equation}
\begin{equation} \label{eq2}
AGB_{musacea} = 0.030 * DBH^{2.13} 
\end{equation}
\begin{equation} \label{eq3}
AGB_{cacao} = 0.1208 * DBH^{1.98} 
\end{equation}
\begin{equation} \label{eq4}
AGB_{timber} = 21.3 - 6.95*DBH + 0.74*DBH^{2}
\end{equation}

\begin{figure}[ht]
\vskip 0.2in
\begin{center}
\centerline{\includegraphics[width=0.9\columnwidth]{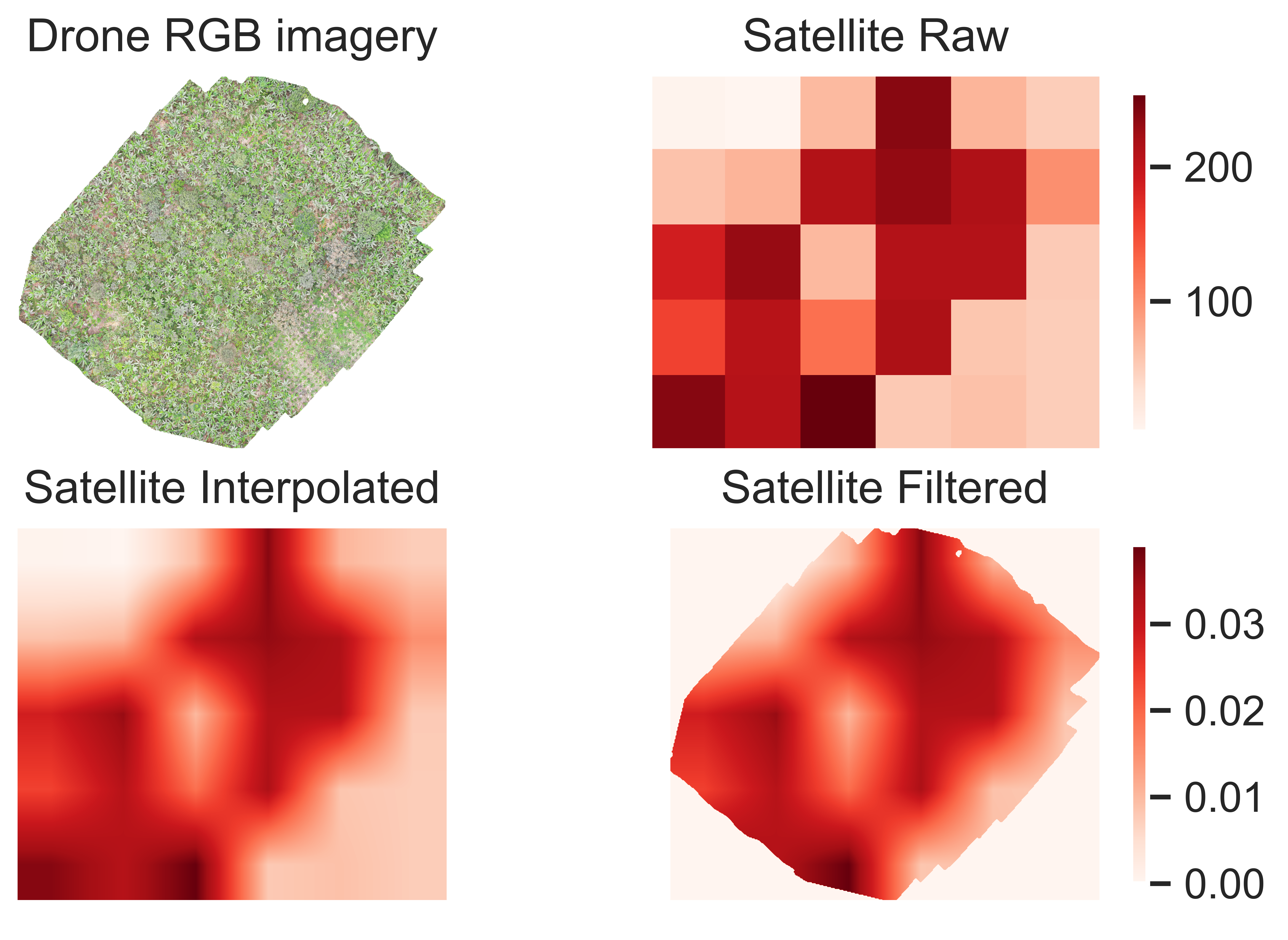}}
\caption{The drone imagery of project site no. 3 with the respective AGB density map from Global Forest Watch.}
\label{figure1}
\end{center}
\vskip -0.2in
\end{figure}

We used the Global Forest Watch (GFW)'s aboveground live woody biomass density dataset as a comparison~\cite{GWF}. It is a global map of AGB and carbon density at 30m x 30m resolution for the year 2000. It is based on more than 700.000 quality-filtered Geoscience Laser Altimeter System (GLAS) lidar observations using machine learning models based on allometric equations for the different regions and vegetation types.

\subsection{Approach}

For each site, we computed the total AGB from the field measurements and the allometric equations. We compare this ground truth with the estimate obtained from GFW for the same location.
\begin{itemize}
  \item Ground Truth: Total AGB values from field measurements divided by the area of the site.
  \item Filtered: The AGB density cubically interpolated to the resolution of the drone imagery and filtered on its polygon. 
\end{itemize}

Comparing the AGB density estimations in tonnes AGB per hectare for each site in Table \ref{result-table} we see that for all plots the satellite-based estimates significantly overestimate the AGB density in the plots, despite their relatively high resolution of 30m x 30m. Drone imagery (1cm/px) combined with convolutional neural networks (CNN) have previously been used to directly estimate biomass and carbon stock in individual trees \cite{jones2020} or indirectly by detecting species or tree metrics such as DBH or H \cite{naafalt18} \cite{schiefer20}, achieving an accuracy similar to manual field measurements. 
We propose an end-to-end carbon stock estimation at the individual tree level by leveraging multi-fusion approaches \cite{Du2020} \cite{Zhang2010} (e.g. combining low-resolution satellite, high-resolution drone imagery, and field measurements or contextual data) and multi-task learning \cite{Crawshaw2020} (e.g. tree metrics and carbon storage factors as auxiliary tasks). 

\section{Conclusion}
There is great potential in combining remote sensing and ML to increase the quality of MRV of forestry carbon offsets and to play a key role in is scaling natural carbon sequestration at the speed necessary to mitigate climate change. However, in this proposal, we identify and highlight the need to audit the algorithms and data used to avoid systematic wrong estimations by quantifying its current gap. We propose to leverage current advancements in remote sensing and ML when creating new automated carbon offset certification protocols, starting with high-resolution data combined with field measurements as benchmarks. 

\subsubsection*{Acknowledgments}
The authors are thankful for the guidance and advice by academic collaborators (Prof. Tom Crowther, Prof. Dava Newman, Simeon Max, Kenza Amara), non-governmental institutions (WWF Switzerland, Restor), and support from the local community in Ecuador.

\bibliography{main}
\bibliographystyle{icml2021}

\clearpage
\appendix

\section{Appendix}
\begin{figure}[ht]
  \centering
  \includegraphics[width=0.50\textwidth]{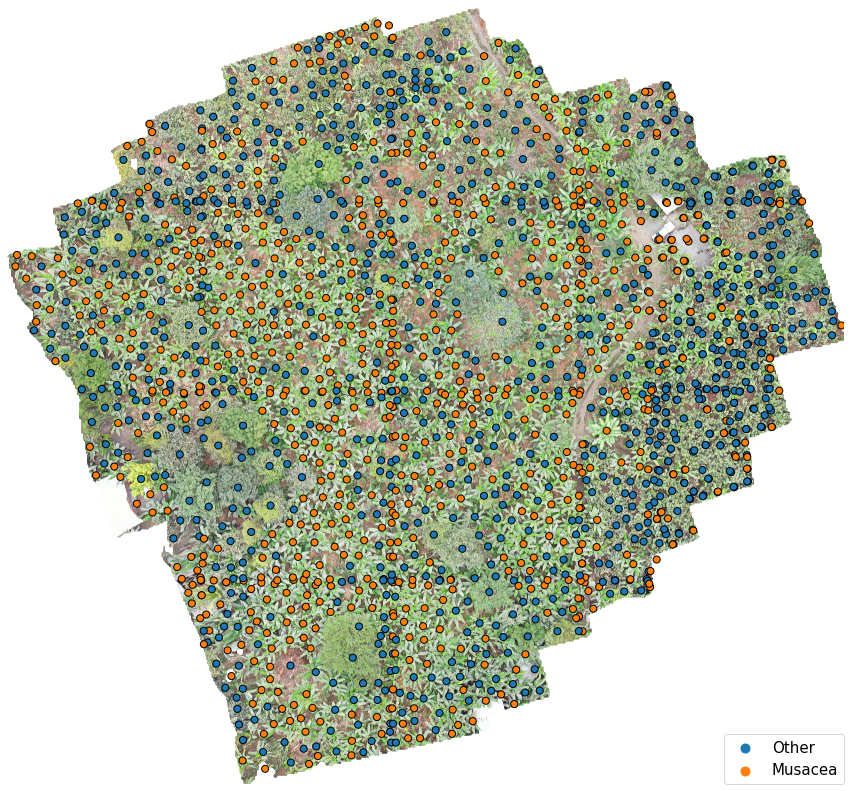}
  \caption{High-resolution aerial mapping and field measurements of a project site. First, we manually collected tree diameter, GPS information and species data for all trees within each of the six project size areas. Afterwards, we ran DeepForest \cite{weinstein2019individual} to detect individual trees and map each field measurement to its corresponding tree. This allows us to apply species-specific allometric equations on a fine-grained resolution and create our ground truth data.}
  \label{fig:appendix}
\end{figure}

\end{document}